\documentclass[
]{ceurart}
\usepackage{caption}
\usepackage{subcaption}
\usepackage{amsfonts}
\usepackage{booktabs}
\usepackage{multirow}
\usepackage{xcolor}
\usepackage[normalem]{ulem} 

\begin{document}

\copyrightyear{2021}
\copyrightclause{Forum for Information Retrieval Evaluation, December 13-17, 2021, India.}

\conference{FIRE 2021: Forum for Information Retrieval Evaluation, 13th-17th December, 2021}

\title{Deep Multi-Task Models for Misogyny Identification and Categorization on Arabic Social Media}


\author[1]{Abdelkader {El Mahdaouy}}[%
orcid=0000-0003-4281-2472,
email=abdelkader.elmahdaouy@um6p.ma,
]

\author[1]{Abdellah {El Mekki}}[%
orcid=0000-0002-7394-3611,
email=abdellah.elmekki@um6p.ma,
]

\author[2]{Ahmed {Oumar}}[%
email=ahmedmohamedlemine.oumar@edu.uca.ma
]

\author[2]{Hajar Mousannif}[
orcid=0000-0002-1307-4215,
email=mousannif@uca.ac.ma
]

\author[1]{Ismail Berrada}[%
orcid=0000-0003-4225-911X,
email=ismail.berrada@um6p.ma,
]
\address[1]{School of Computer Sciences, Mohammed VI Polytechnic University, Morocco}
\address[2]{LISI Laboratory, Computer Science Department, FSSM, Cadi Ayyad University, Morocco}

\begin{abstract}
 The prevalence of toxic content on social media platforms,
 such as hate speech, offensive language, and misogyny, presents serious challenges to our interconnected society. These challenging issues have attracted widespread attention in Natural Language Processing (NLP) community. In this paper, we present the submitted systems to the first Arabic Misogyny Identification shared task. We investigate three multi-task learning models as well as their single-task counterparts. In order to encode the input text, our models rely on the pre-trained MARBERT language model. The overall obtained results show that all our submitted models have achieved the best performances (top three ranked submissions) in both misogyny identification and categorization tasks. 
\end{abstract}


\maketitle
\section{Introduction}
With the popularity of the Internet and the rise of social media platforms, users around the world are having more freedom of expression. They can express their thoughts and opinions with minimal limitations and restrictions. As a result, they can share their positive thoughts about a specific product or service, a political decision, etc. Besides, they can share their negative thoughts about other things. Unfortunately, many users can employ these communication channels and freedom of expression to bully other people or groups. Misogyny is one of these phenomena, and it is defined as hate speech towards the female gender \cite{Moloney2018AssessingOM}. Misogyny can be classified into several categories such as sexual harassment, damning, dominance, etc \cite{bookhaters}.

Misogynistic behavior has prevailed on social media such as Facebook and Twitter. The ease of use and richness of these platforms have upraised misogyny to new levels of violence around the globe. Moreover, women suffer from misogyny in the 1st tier world as they suffer from it in the 2nd and 3rd tier world despite their race, language, age, etc.  In the Arabic world, women's rights and liberty have been always a controversial subject. Therefore, women are also exposed to online misogyny, where people can start campaigns of intimidation and harassment against them for one reason or another.

Fighting online misogyny has become a topic of interest of several Internet players, where social media networks such as Facebook and Twitter propose reporting systems that allow users to post messages expressing misogynistic behavior. These reporting systems can detect these behaviors from users' posts and delete them automatically. For high-resource languages such as English, Spanish, and French, these systems have been shown to perform well. However, when it comes to languages such as Arabic, automatic reporting systems are not yet deployed, and that is mainly due to: 1) the lack of annotated data needed to build such systems and 2) the complexity of the Arabic language compared to other languages.

Fine-tuning pre-trained transformer-based language models \cite{devlin-etal-2019-bert} on downstream tasks has shown state-of-the-the-art (SOTA) performances on various languages including Arabic \cite{antoun-etal-2020-arabert, abdul-mageed-etal-2021-arbert, el-mekki-etal-2021-domain, el-mekki-etal-2021-bert, el-mahdaouy-etal-2021-deep}. Although several research works based on pre-trained transformers have been introduced for misogyny detection in Indo-European languages \cite{safi-samghabadi-etal-2020-aggression, FersiniNR20,9281090}, works on Arabic language remain under explored \cite{mulki2021letmi}.

In this paper, we present our participating system and submissions to the first Arabic Misogyny Identification (ArMI) shared tasks~\cite{armi2021overview}. We introduce three Multi-Task Learning (MTL) models and their single-task counterparts. To embed the input texts, our models employ the pre-trained MARBERT language model \cite{abdul-mageed-etal-2021-arbert}. Moreover, for Task 2, we tackle the class imbalance problem by training our models to minimize the Focal Loss \cite{abs-1708-02002}. The obtained results demonstrate that our three submissions have achieved the best performances for both ArMI tasks in comparison to the other participating systems. The results also show that MTL models outperform their single-task counterparts on most evaluation measures. Additionally, the Focal Loss has shown effective performances, especially on F1 measures.

The rest of this paper is organized as follows. Section \ref{sec:armi} describes the ArMI tasks and the provided dataset. In Section \ref{sec:system}, we introduce our participating system and the investigated deep learning models. Section \ref{sec:exps} presents the conducted experiments and shows the obtained results. In section \ref{sec:conc}, we conclude the paper. 

\section{Tasks and dataset description}
\label{sec:armi}
\begin{figure*}[ht]
     \centering
     
     \begin{subfigure}[b]{0.40\textwidth}
         \centering
    \includegraphics[scale=0.5]{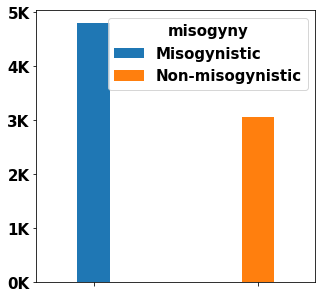}
    \caption{Distribution of misogynistic tweets}
    \label{fig:binary_dist}
     \end{subfigure}
     \hfill
     \begin{subfigure}[b]{0.55\textwidth}
         \centering
    \includegraphics[scale=0.5]{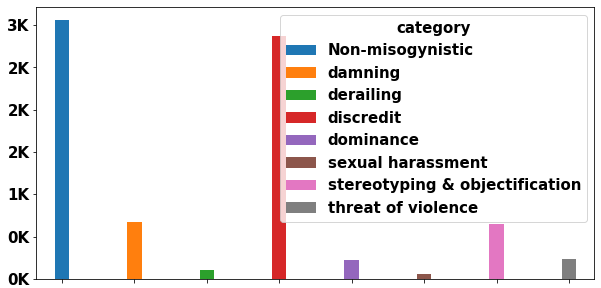}
    \caption{Distribution of misogynistic categories}
    \label{fig:multi_dist}
    \end{subfigure}

\caption{Labels distribution for both misogyny and category detection tasks. \label{fig:armidata}}       
\end{figure*}

The Arabic Misogyny Identification (ArMI) task consists of the automatic detection of misogyny from Arabic tweets~\cite{armi2021overview}. This task is composed of two main sub-tasks: the 1st sub-task is a binary classification task where the objective is to classify whether a tweet is misogynistic or not. In the second sub-task, the objective is to detect the misogynistic behavior expressed in a tweet. It is modeled as a multi-class classification problem consisting of seven misogynistic behaviors (labels). The organizers of this task have provided 7,866 labeled tweets to serve both sub-tasks for model training, while 1,966 tweets have been used for model testing and evaluation. Figure \ref{fig:armidata} presents the distribution of both tasks labels. It shows that the class labels are imbalanced for both misogyny identification and categorization tasks. 

The provided tweets are expressed mainly in Modern Standard Arabic (MSA), while several tweets are expressed in some Arabic dialects such as Egyptian, Gulf, and Levantine. The Levantine tweets are taken from Let-Mi misogyny detection dataset, proposed by \citet{mulki2021letmi}. Besides, the rest of the tweets have been scrapped from Twitter using hashtags related to the misogyny phenomenon. The provided dataset is manually annotated by Arabic native speakers.

\section{Methodology}
\label{sec:system}
We propose three deep Multi-task Learning (MTL) models based on the pre-trained MARBERT encoder \cite{abdul-mageed-etal-2021-arbert} for the ArMI shared task. We also investigate the single-task version of the proposed MTL models. The choice of MARBERT encoder is motivated by the fact that this language model is pre-trained on 1B tweet corpus, containing both dialectal Arabic and MSA. Moreover, Fine-tuning MARBERT on downstream NLP tasks has shown effective results in many Arabic NLP applications \cite{abdul-mageed-etal-2021-arbert, el-mekki-etal-2021-bert, el-mahdaouy-etal-2021-deep}. In what follows, we describe each component of our submitted system.

\subsection{Preprocessing}

The tweet preprocessing component performs emojis extraction, user mention and URL substitution, and hashtag normalization. Following MARBERT's tweets preprocessing guidelines, user mentions and URLs are replaced by "user" and "url" token, respectively. For hashtags normalization, we remove "\#" symbol and replace "\_" by white space. It is worth mentioning that diacritics are already removed from the training and testing datasets. Based on our preliminary experiments, emojis are not removed from the normalized text and added after the [SEP] token of the employed encoder. Finally, each tweet is represented using its normalized text and its emojis, as follows: 
\begin{itemize}
    \item[$\star$] [CLS] normalized tweet [SEP] emojis [SEP]
\end{itemize}

\subsection{Deep Learning Models}

In this section, we describe the employed MTL models and their single task counterparts. All our models utilize MARBERT encoder to represent the input tweets. The models are described as follows: 
\begin{itemize}
    \item \textbf{MT\_CLS} uses a classification layer for each task on top of MARBERT encoder. It relies on [CLS] token embedding to predict the class label for each task. The single-task version of this model is denoted by \textbf{ST\_CLS}.  
    \item \textbf{MT\_ATT} consists of MARBERT encoder, two task-specific attention layers, and two classification layers. Each attention layer \cite{Bahdanau2015NeuralMT,yang-etal-2016-hierarchical} extracts task discriminative features by weighting the output token embedding of the encoder according to their contribution to the task at hand. Each classification layer is feed with the concatenation of the task attention output and the [CLS] token embedding. This model has shown effective performances in many NLP tasks, including dialect identification, sentiment analysis and sarcasm detection for the Arabic language \cite{el-mekki-etal-2021-bert, el-mahdaouy-etal-2021-deep}, humor detection and rating, as well as lexical complexity prediction in English \cite{ essefar-etal-2021-cs, el-mamoun-etal-2021-cs}. The single-task counterpart of \textbf{MT\_ATT} is denoted by \textbf{ST\_ATT}.
    
    \item \textbf{MT\_VHATT} is an extension of the \textbf{MT\_ATT} model. In addition to the task-specific attention layers (called horizontal attention layers), it employs vertical attention layers to incorporate the features of the top intermediate layers of MARBERT encoder for both tasks. This model utilizes six attention layers to extract features from the token embedding of the top six layers of the encoder \cite{Bahdanau2015NeuralMT,yang-etal-2016-hierarchical}. Then, another attention layer is employed to aggregate features from the six vertical attention layers. Note that, we exclude the top output layer of the encoder as its features are already used by the horizontal attention layers (task-specific attention). Finally, the input of the classification layers for both tasks is the concatenation of the [CLS] token embedding of the last layer of the encoder, the task-specific attention output, and the aggregated features of intermediate layers. The single-task version of this model (\textbf{MT\_VHATT}) is denoted by \textbf{ST\_VHATT}.            
\end{itemize}

For misogyny identification (Task 1), all models are trained to minimize the binary cross-entropy loss. For misogyny categorization (Task 2), we have investigated the Cross-Entropy (CE) loss, as well as the Focal Loss (FL) \cite{abs-1708-02002}. The latter loss is employed to handle the class imbalance problem. It reduces the loss contribution from easy examples and assigns higher importance weights for hard-to-classify examples. The FL is given by:

\begin{equation}
    FL(y, \hat{p})
        = - \alpha_y \left(1 - \hat{p}_y\right)^\gamma \log(\hat{p}_y)
\end{equation}
where, $y \in \{0, \ldots, K - 1\}$ denotes the category's label, $\hat{p} = (\hat{p}_0, \ldots, \hat{p}_{K-1})$ is a vector representing the predicted probability distribution over the labels, $\alpha_y$ is the weight of label $y$, and $\gamma$ controls the contribution of high-confidence predictions in the loss. In other words, a higher value of $\gamma$ implies lower loss contribution for well-classified examples \cite{abs-1708-02002}. 

\section{Experiments and results}
\label{sec:exps}
In this section, we present the experiment settings as well as the obtained results for our development set and the provided test set. 

\subsection{Experiment settings}

All our models are implemented using PyTorch\footnote{\url{https://pytorch.org/}} framework and the open-source Transformers\footnote{\url{https://huggingface.co/transformers/}} libraries. Experiments are performed using a PowerEdge R740 Server, having 44 cores Intel Xeon Gold 6152 2.1GHz, a RAM of 384 GB, and a single Nvidia Tesla V100 with 16GB of RAM. The provided training set is split into $90\%$ for the training and $10\%$ for the development. Based on our preliminary results, all models are trained using Adam optimizer. The learning rate, the number of epochs, and the batch size are fixed to $1 \times 10^{-5}$, $5$, and $16$ respectively. The hyper-parameter $\gamma$ of the Focal Loss is set to $2$, while the weights of Task 2 labels are set to $\alpha_y = \frac{\text{number of instance of dominant label}}{\text{number of instance of label y}}$. All models are evaluated using the Accuracy as well as the macro averaged Precision, Recall, and F1 measures. 

\subsection{Results}
In order to select the best models for our official submissions, we have evaluated the three MTL models and their single-task counterparts. For Task 2, we have investigated both CE and FL losses. Table \ref{tab:stdev} presents the obtained results on the development set using the three single-task models. The overall obtained results for Task 1 show that the ST\_ATT model outperforms the other models on most evaluation measures. It shows also the best Recall and F1 measures for Task 2. Moreover, ST\_VHATT yields slightly better performances on Task 1 and achieves far better precision and F1 scores on Task 2 in comparison to ST\_CLS model. Furthermore, FL outperforms the CE loss on most evaluation measures for Task 2, except for the accuracy and the precision of model ST\_CLS. Table \ref{tab:cefl} presents the classification reports for Task 2 of the ST\_ATT model using CE and FL loss functions. The obtained results show that the FL leads to better F1 scores for all categories, except "Discredit" and "Damning" misogynistic behaviours. Indeed, the classification of rare events is increased while maintaining the overall performance.

\begin{table}[htbp]
  \centering
  \caption{The obtained results on the dev set using the three single-task models for both Task 1 and Task 2.}
  \resizebox{1\textwidth}{!}{%
    \begin{tabular}{lcccc|ccccc}
\cmidrule{2-10}          & \multicolumn{4}{c|}{Task 1}   & \multicolumn{5}{c}{Task 2} \\
    \midrule
    Model & Accuracy & Precision & Recall & F1    & Cat. Task Loss & Accuracy & Precision & Recall & F1 \\
    \midrule
    \multirow{2}[2]{*}{ST\_CLS} & \multirow{2}[2]{*}{90.72} & \multirow{2}[2]{*}{90.48} & \multirow{2}[2]{*}{89.91} & \multirow{2}[2]{*}{90.18} & CE    & \textbf{81.58} & \textbf{71.80} & 56.05 & 60.66 \\
          &       &       &       &       & FL    & 79.67 & 62.15 & 63.00 & 62.05 \\
    \midrule
    \multirow{2}[2]{*}{ST\_ATT} & \multirow{2}[2]{*}{\textbf{90.98}} & \multirow{2}[2]{*}{\textbf{90.80}} & \multirow{2}[2]{*}{90.12} & \multirow{2}[2]{*}{\textbf{90.43}} & CE    & 80.81 & 67.79 & 56.70 & 59.63 \\
          &       &       &       &       & FL    & 80.94 & 70.22 & \textbf{62.12} & \textbf{64.60} \\
    \midrule
    \multirow{2}[2]{*}{ST\_VHATT} & \multirow{2}[2]{*}{90.85} & \multirow{2}[2]{*}{90.50} & \multirow{2}[2]{*}{\textbf{90.20}} & \multirow{2}[2]{*}{90.34} & CE    & 80.43 & 64.87 & 60.15 & 61.99 \\
          &       &       &       &       & FL    & 80.94 & 68.43 & 61.79 & 63.96 \\
    \bottomrule
    \end{tabular}%
    }
  \label{tab:stdev}%
\end{table}%

\begin{table}[htbp]
  \centering
  \caption{ST\_ATT model's classification reports on the dev set of Task 2 using CE and FL loss functions.}
  \resizebox{1\textwidth}{!}{%
    \begin{tabular}{l|ccc|ccc|c}
    \toprule
    \multirow{2}[4]{*}{Category} & \multicolumn{3}{c|}{CE loss} & \multicolumn{3}{c|}{FL loss} & \multirow{2}[4]{*}{Support} \\
\cmidrule{2-7}          & Precision & Recall & F1    & Precision & Recall & F1    &  \\
    \midrule
    None  & 0.8509 & 0.8954 & 0.8726 & 0.8845 & 0.8758 & 0.8801 & 306 \\
    Damning & 0.8841 & 0.9104 & 0.8971 & 0.8955 & 0.8955 & 0.8955 & 67 \\
    Derailing & 0.2500 & 0.0909 & 0.1333 & 0.4286 & 0.2727 & 0.3333 & 11 \\
    Discredit & 0.8247 & 0.8362 & 0.8304 & 0.7980 & 0.8397 & 0.8183 & 287 \\
    Dominance & 0.3636 & 0.3636 & 0.3636 & 0.4375 & 0.3182 & 0.3684 & 22 \\
    Sexual harassment & 1.0000 & 0.3333 & 0.5000 & 1.0000 & 0.5000 & 0.6667 & 6 \\
    Stereotyping \& objectification & 0.6786 & 0.5846 & 0.6281 & 0.6897 & 0.6154 & 0.6504 & 65 \\
    Threat of violence & 0.5714 & 0.5217 & 0.5455 & 0.4839 & 0.6522 & 0.5556 & 23 \\
    \bottomrule
    \end{tabular}%
    }%
  \label{tab:cefl}%
\end{table}%

\begin{table}[htbp]
  \centering
  \caption{The obtained results on the dev set using the three multi-task models for both Task 1 and Task 2.}
    \resizebox{1\textwidth}{!}{%
    \begin{tabular}{l|c|llll|llll}
    \cmidrule{3-10}
          \multicolumn{1}{c}{}&    \multicolumn{1}{c}{}   & \multicolumn{4}{c|}{Task 1} & \multicolumn{4}{c}{Task 2} \\
    \midrule
    Model & Cat. Task Loss & Accuracy & Precision & Recall & F1    & Accuracy & Precision & Recall & F1 \\
    \midrule
    \multirow{2}[1]{*}{MT\_CLS} & CE    & 90.98  & 90.39 & 90.72 & 90.55 & 79.67 & 67.68 & 57.02 & 60.18 \\
          & FL    & 91.49 & 91.34 & 90.66 & 90.97 & 80.43 & 67.65 & 60.55 & 62.92 \\
    \midrule
    \multirow{2}[2]{*}{MT\_ATT} & CE    & 91.11 & \textbf{91.48} & 89.75 & 90.46 & 80.56 & 66.80 & 58.63 & 60.52 \\
          & FL    & \textbf{91.74} & 91.42 & \textbf{91.16} & \textbf{91.28} & \textbf{80.81} & \textbf{67.90} & \textbf{61.55} & \textbf{63.29} \\
    \midrule
    \multirow{2}[2]{*}{MT\_VHATT} & CE    & 91.11 & 90.81 & 90.40 & 90.60 & 80.18 & 66.91 & 57.39 & 60.01 \\
          & FL    & 91.49 & 91.19 & 90.84 & 91.01 & 80.05 & 66.82 & 58.92 & 61.67 \\
    \bottomrule
    \end{tabular}%
    }%
  \label{tab:mtdev}%
\end{table}%

Table \ref{tab:mtdev} presents the obtained results on the dev set using the three multi-task models for both Task 1 and Task 2. The overall obtained results show that the MT\_ATT outperforms all other models on both tasks for most evaluation measures. The results demonstrate that using the FL loss for Task 2 improves also the model's performance on Task 1 in multi-task settings. In accordance with the obtained results using single-task models, MT\_VHATT shows slightly better performances on Task 1 than ST\_CLS model. The overall obtained show that muti-task learning models surpass their single-task counterparts on Task 1. This can be explained by the fact MT models leverage signals from both tasks \cite{Caruana94,sun2019ernie20}.       

\subsection{Official submissions results}
Based on the obtained results on the development set, we have submitted models that are trained using the FL for misogyny categorization (Task 2). This choice is motivated by the fact that the FL loss has lead to better F1 scores (Tables \ref{tab:stdev} and \ref{tab:mtdev}) than CE loss on the dev set. Our three official submissions are described as follows:
\begin{itemize}
    \item \textbf{run1}: corresponds to the submission of the obtained results on both tasks using the single-task model \textbf{ST\_ATT}.  
    \item \textbf{run2}: corresponds to the obtained results on both tasks using the multi-task model \textbf{MT\_ATT}.
    \item \textbf{run3}: corresponds to the ensembling of the three multi-task learning models, namely \textbf{MT\_CLS}, \textbf{MT\_ATT}, and \textbf{MT\_VHATT} models. In this submission, the logits of the three models are averaged. Depending on the task, either the sigmoid or the softmax activation is applied to get the labels probabilities.   
\end{itemize}

\begin{table}[htbp]
\caption{Top five submitted systems's performance on ArMI Task 1.}
  \resizebox{0.6\textwidth}{!}{%
    \begin{tabular}{l|cccc}
\cmidrule{2-5}    \multicolumn{1}{r}{} & Accuracy & Precision & Recall & F1 \\
    \midrule
    UM6P-NLP\_run3 & \textbf{91.9} & \textbf{92}  & 90.9 & \textbf{91.4} \\
    UM6P-NLP\_run2 & 91.5 & 91.5 & 90.5 & 91 \\
    UM6P-NLP\_run1 & 91.5 & 91.1 & \textbf{91.1} & 91.1 \\
    \midrule
    UoT\_run1 & 90.5 & 90.1 & 89.9 & 90 \\
    SOA\_NLP\_run1 & 88.3 & 87.8 & 87.6 & 87.7 \\
    \bottomrule
    \end{tabular}%
    \label{tab:task1}%
    }
\end{table}  

\begin{table}[htbp]
  \caption{Top five submitted systems's performance on ArMI Task 2.}
  \resizebox{.6\textwidth}{!}{%
    \begin{tabular}{l|cccc}
\cmidrule{2-5}    \multicolumn{1}{r}{} & Accuracy & Precision & Recall & F1 \\
    \midrule
    UM6P-NLP\_run2 & 82.7 & 69.7 & 64.7 & \textbf{66.5} \\
    UM6P-NLP\_run3 & \textbf{83.3} & \textbf{71.7} & 63.6 & 65.3 \\
    UM6P-NLP\_run1 & 81.6 & 69.2 & \textbf{65.2} & 65.1 \\
    \midrule
    SOA\_NLP\_run2 & 76.4 & 67.6 & 48  & 53.1 \\
 SOA\_NLP\_run3 & 74.5 & 54.9 & 50.8 & 52.6 \\
    \bottomrule
    \end{tabular}%
    }
  \label{tab:task2}%
\end{table}
Tables \ref{tab:task1} and \ref{tab:task2} summaries the official results of top five submitted systems to Task 1 and Task 2 respectively. The results show that all our submissions are ranked top three among all submitted systems. In accordance with our previous results, our multi-task models have achieved the first and the second-ranking positions. Although the ensembling of the three MTL models (run3)  has yielded the best performances on most evaluation measures for both tasks, the best F1 score for Task 2 is obtained by run2 (MT\_ATT model).

\section{Conclusion}
\label{sec:conc}
In this paper, we have presented our participating system in the first Arabic Misogyny Identification shared task. We have investigated three Multi-Task Learning models and their single-task counterparts using the pre-trained MARBERT encoder. In order to deal with class labels imbalance for Task 2, we have employed the Focal Loss. The results show that our three submitted systems are top-ranked among the participating systems to both ArMI tasks. The overall obtained results demonstrate that MTL models outperform their single-task versions in most evaluation scenarios. Besides, the Focal Loss has shown effective performances, especially on F1 measures.   
\begin{acknowledgments}
Experiments presented in this paper were carried out using the supercomputer simlab-cluster, supported by Mohammed VI Polytechnic University (\url{https://www.um6p.ma}), and facilities of simlab-cluster HPC \& IA platform.
\end{acknowledgments}

\bibliography{sample-ceur}


\end{document}